\documentclass{sig-alternate}

\usepackage{definitions}
\usepackage{graphicx}
\usepackage{subcaption}

\usepackage[ruled]{algorithm2e}

\SetAlFnt{\small}
\SetAlCapFnt{\small}
\SetAlCapNameFnt{\small}
\SetAlCapHSkip{0pt}
\begin{document}
%

\title{Causal Discovery for Manufacturing Domains}

\numberofauthors{4} 
%
\author{
\alignauthor
Katerina Marazopoulou\\
       \affaddr{College of Information and Computer Sciences}\\
       \affaddr{UMass Amherst}\\
       \email{kmarazo@cs.umass.edu}
\alignauthor
Rumi Ghosh\\
      \affaddr{Data Mining Services and Solutions, Robert Bosch LLC}\\
      \affaddr{Palo Alto, USA}\\
      \email{rumi.ghosh@us.bosch.com}
\alignauthor Prasanth Lade\\
      \affaddr{Data Mining Services and Solutions, Robert Bosch LLC}\\
      \affaddr{Palo Alto, USA}\\
      \email{prasanth.lade@us.bosch.com}
\and
\alignauthor David Jensen\\
      \affaddr{College of Information and Computer Sciences}\\
       \affaddr{UMass Amherst}\\
       \email{jensen@cs.umass.edu}
}

\maketitle

\begin{abstract}

Increasing yield and improving quality are of paramount importance to any manufacturing company. One of the ways to achieve this is through discovery of the causal factors that affect these quantities. In this work, we use data-driven causal models to identify causal relationships in manufacturing. Specifically, we apply causal structure learning techniques on real data collected from a production line. Emphasis is given to the interpretability of the learned causal models, so that they can be used by practitioners to take meaningful actions. We highlight the challenges presented by assembly-line data and propose ways to address those challenges. We also identify unique characteristics of data originating from assembly lines and how to leverage those characteristics to improve causal discovery.  Standard evaluation techniques for causal structure learning show that the learned models closely match the underlying causal relationships between different factors in the production process. These results were also validated by manufacturing domain experts, who found them promising. This work demonstrates how data mining and knowledge discovery can be used for root cause analysis in the domain of manufacturing and connected industry.

\end{abstract}


\category{H.2.8}{Database Management}{Database Applications}[Data mining]
\category{I.2.6}{Artificial Intelligence}{Learning}[Concept learning, knowledge acquisition]
\category{J.1}{Computer Applications}{Administrative Data Processing}[Manufacturing]
\keywords{Causal Discovery, Manufacturing, Structure Learning}

\section{Introduction}
With increasing interest in the Internet of Things (IoT), the size and complexity of data sets collected from manufacturing processes has grown significantly.
The ability to use networked devices and sensors in assembly lines has enabled collection of data from every stage of the manufacturing process. However, the collection of arbitrarily large amounts of data is meaningful only if it leads to actionable insights, especially on how to increase yield and efficiency. Our work is a step in this direction.

This work has two objectives.
First, we aim to identify the joint causal structure of a specific manufacturing domain. 
Second, we focus on the the key causal factors that increase yield. 
To be more specific, given the measurements taken during the production and testing of a product, we aim to identify the causal relationships of the domain, which include the influential factors affecting yield. 

The traditional approach to address this issue is through Design of Experiments (DoE)~\cite{fisher1960design}. 
In the context of manufacturing, DoE usually includes carrying out experiments in the real production and testing environment. As a consequence, it can be costly and time-consuming.
Therefore, practitioners limit the number of factors examined in each experiment. 
As the process of manufacturing becomes more complex, sometimes comprising hundreds of possible factors, it becomes very difficult to choose the appropriate factors for further investigation through DoE. 
Currently, most experts in manufacturing rely on domain knowledge and intuition along with basic statistics to guide their efforts to increase yield and improve quality. 

This work aims to aid this process 
by using data mining and knowledge discovery. 
Specifically, we apply techniques for learning causal structure from real data collected  from a manufacturing line in order to identify the key factors affecting yield and the complex interrelationships amongst them\footnotemark{}.
\footnotetext{Efforts are ongoing to make this data set public for the benefit of the knowledge discovery and data mining communities.}

Algorithms for learning causal structure construct a causal graphical model over the variables of a domain.
Such models aim to identify the underlying causal mechanisms of a specific domain. 
Causal models are of great value, because knowing the causes of a target variable specifies interventions on other variables that can change the target variable in a desired way.
Causal models have been applied in areas such as planning, decision making, epidemiology, and social science, just to mention a few.

In what follows, we first describe the manufacturing domain, i.e., the manufacturing process and the structure of an assembly line.
We then provide a short introduction to graphical models and causal discovery algorithms.
We show qualitative results obtained from application of causal discovery algorithms on the data collected from a production line.
We identify the specific challenges presented by manufacturing data and how the standard algorithms need to be modified in order to improve causal discovery.
Finally, we present an evaluation of causal structure learning for manufacturing, based on domain expertise and synthetic data.

\section{The Manufacturing Domain}

 Typical assembly lines consist of multiple stations where different operations take place. In every station, several measurements are taken for the product up to that point. Components are added to an unfinished product in different  \emph{production stations} of an assembly line. A \emph{testing station} is a station where a product passing through is inspected. Additionally, at the end of the assembly line, there are usually testing stations---called \emph{end-of-line (EOL)} testing stations---which inspect the final product. 
 Testing stations carry out a series of measurements to test the quality of the product. If the product does not meet the required quality criteria, it is usually rejected. A rejected product is called a \emph{scrap} and an accepted product is called a \emph{good part}.
 
\begin{figure}[t]
\centering
\includegraphics[width=\columnwidth]{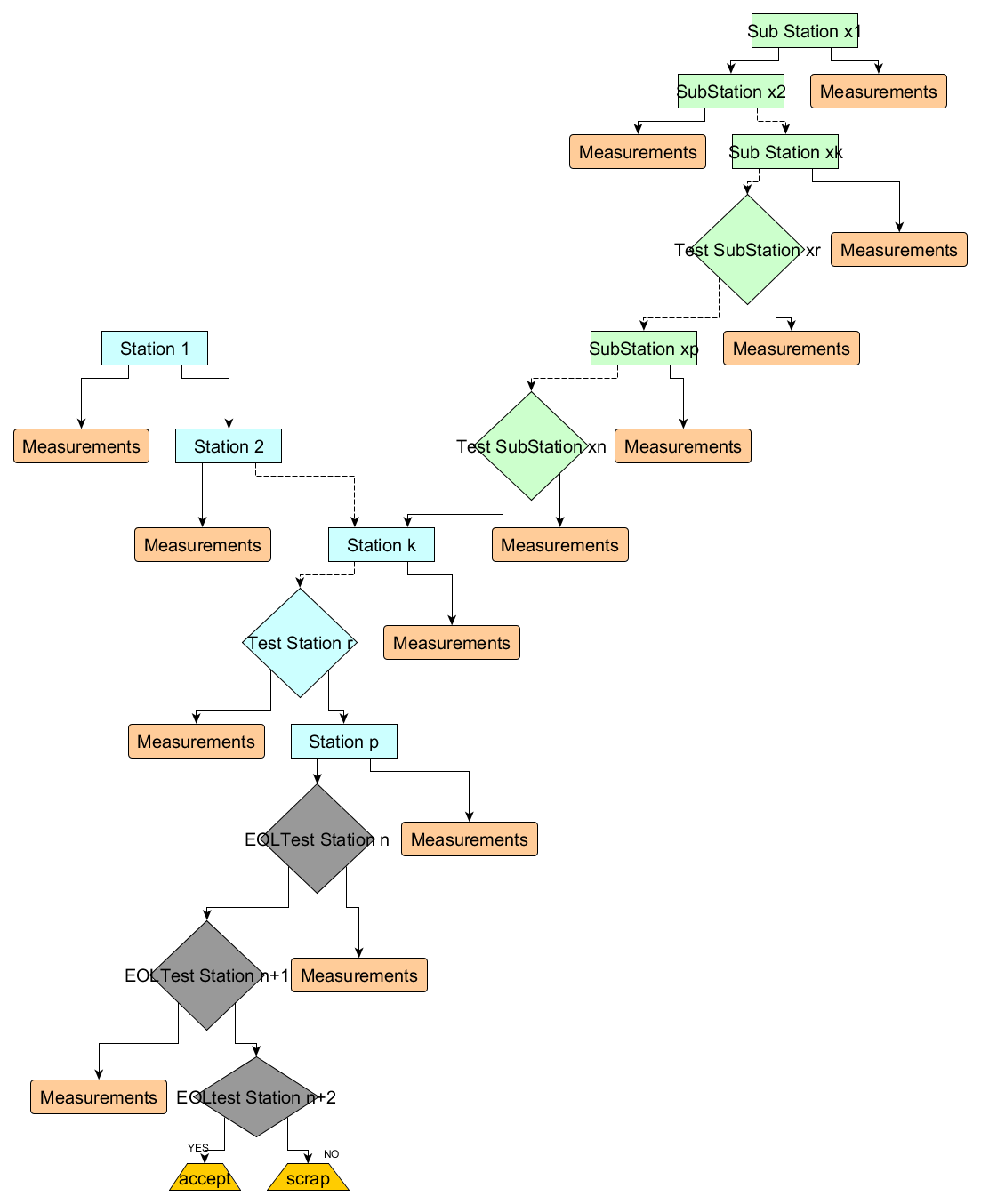}
\caption{A schematic representation of an assembly line.
Blue denotes stations in the main assembly line, green stations in assembly lines of sub-components, grey denotes EoL stations, orange denote measurements, and yellow is the output of the process.
Rectangles (blue and green) denote production stations, rhombi denote testing stations.
}
\label{fig:AssemblyLine}
\end{figure}

 An illustration of an assembly line is shown in Figure~\ref{fig:AssemblyLine}. Production stations are depicted with a rectangle and testing stations with a rhombus. The EOL testing stations are shown in gray. For example, in Figure~\ref{fig:AssemblyLine}, stations 1, 2, k and p are production stations depicted by blue rectangles. Station r is a test station represented by a blue rhombus. The measurements collected in the different stages of the production cycle are shown in orange rectangles.
 
The components that are used in the main assembly line to form the finished product are also assembled separately in assembly lines. For example, the substations in green in Figure~\ref{fig:AssemblyLine} show the assembly line for the component that is added to the main product in \emph{station k}. In the era of the Internet of Things and connected industry, the  information  gathered from the measurements in the main assembly line can be augmented with the information from the assembly lines for the individual sub-components. Measurements are also gathered externally, from suppliers of sub-components. We use the term \emph{manufacturing cycle data} to cover data gathered from the assembly line of the component, as well as from the assembly lines of the sub-components and the supplier data. 

In this work, the  product under investigation is a fuel injector. 
Measurements were recorded  from the assembly line producing the injector and from assembly lines producing the sub-components of the injector. 
Our goal is to leverage this plethora of measurements to learn the complex causal relationships between factors affecting yield in the production process. 
However, the methodology for root cause analysis proposed in this paper is generic and can easily be applied to any product manufactured in an assembly line.

\subsection{Description of the data used}

The raw data produced by the assembly line for injectors consists of over 69,000 factors (or variables) and over one million parts (data points) manufactured within a period of a year. Thus, this is a massive data set from a manufacturing system, both in terms of data points and number of variables measured.
For our analysis, the data was preprocessed to ignore features that are unique keys and features with zero variance.
Moreover, we considered features from a single assembly line, one substation and one testing line. The clean data set consists of 431 continuous variables, normalized to 0 mean and variance 1. 

It is worth mentioning that, even after preprocessing, the variables exhibit high pairwise linear correlation, as shown in Figure~\ref{fig:correlation-original}, and non-linear relationships. These characteristics make the task of causal discovery challenging. 
However, there is a large amount of domain knowledge and inherent structure in the data that can be leveraged to improve the results.

\section{Preliminaries}

In this section we introduce the notation and basic definitions that we will use throughout the paper. Upper-case letters denote random variables (e.g., $X$, $V$). 
Bold-face fonts denote sets of random variables (e.g., $\mathbf{Z}$).  
First, we provide a short introduction to Bayesian networks and their causal interpretation. Then, we describe how to learn the structure of such models from data.

\subsection{Bayesian Networks}

Bayesian networks are directed graphical models that compactly represent sets of probability distributions. 
A \emph{Bayesian network} defined over a set of random variables $\mathbf{V}$ consists of:
\begin{enumerate}
\item A directed acyclic graph $G=\langle \mathbf{V}, \mathbf{E} \rangle$, known as the \emph{structure of the network}, and
\item A set of parameters $\mathbf{\Theta}$, where every parameter $\theta\in\mathbf{\Theta}$ is a conditional distribution of a node given its parents in the graph. 
\end{enumerate}

A graph $G$ is a \emph{partially directed graph} if it has both directed and undirected edges. 
The \emph{skeleton} of a graph $G$ is the undirected graph that can be obtained by substituting every edge of $G$ with an undirected edge. 
A \emph{path} between two nodes $X$ and $Y$ is a sequence of nodes $X, V_1, \ldots, V_n, Y$ such that there exists an edge between every two consecutive nodes of the path. 
A \emph{v-structure} or \emph{collider} on a graph $G$ is an ordered triple of nodes $\langle X, Y, Z\rangle$ such that $X\rightarrow Y\leftarrow Z$ and there is no edge between $X$ and $Z$.

The structure of a Bayesian network represents a set of independencies according to the local Markov condition: a node is independent of its non-descendants given its parents in the graph. 
The same set of independencies can be derived through the use of the graphical criterion of \textit{d}-separation.
We say that two nodes $V_1$ and $V_2$ are \emph{\textit{d}-separated} given a disjoint set of variables $\mathbf{Z}$ if there are no \textit{d}-connecting paths between $V_1$ and $V_2$ given $\mathbf{Z}$. 
A path between $V_1$ and $V_2$ is \emph{d-connecting} given a set $\mathbf{Z}$, if every collider along the path is in $\mathbf{Z}$ or has a descendant in $\mathbf{Z}$ and no other nodes are in $\mathbf{Z}$.
It is worth noting that different directed acyclic graphs might represent the same set of independencies. 
That set of DAGs is known as the \emph{Markov equivalence class} for that set of independencies.
A Markov equivalence class can be graphically represented by a mixed graph (a graph that contains both directed and undirected edges). 
For every undirected edge, any of the two possible orientations does not change the set of conditional independencies induced by the graph.

\subsection{Causal Bayesian Networks}

The notion of causality has been a subject of debate for philosophers and practitioners since ancient times. 
In this work, we focus on the semantics of causality based on probabilistic distributions and manipulations, following the work of Pearl~\cite{pearl2009causality} and Spirtes et al.~\cite{spirtes-etal-book00}.
In this framework, $X$ is a \emph{direct cause} of $Y$ with respect to a set of variables $\mathbf{Z}$ if changing the value of $X$ results in changes in the probability distribution of $Y$, assuming that the values of all other variables in $\mathbf{Z}$ are held constant~\cite{spirtes2010intotocausal}.

To interpret graphs (and thus Bayesian networks) causally and use them for causal inference and discovery, we need to make an additional set of assumptions. 
\begin{enumerate}
\item \emph{Causal Sufficiency:}
There are no unmeasured common causes of a measured variable.
\item \emph{Causal Markov condition:}
The Markov condition and \textit{d}-separation provide a connection between the structure of a Bayesian network and the independencies that hold in the underlying distribution. 
If we want to interpret Bayesian networks causally, a stronger form of the Markov condition is needed, the \emph{causal Markov} condition:

A variable is independent of its non-effects conditioned on its direct causes.

This implies that the independencies that can be read off the structure of the graph ($I_G$) are a subset of the independencies that hold in the underlying distribution ($I_D$): $I_G\subseteq I_D$.
\item \emph{Faithfulness:} 
This assumption states that the set of independencies that hold in the distribution, can be read from the structure of the graph: $I_D\subseteq I_D$.
\end{enumerate}
In a causal Bayesian network, a directed edge $X\rightarrow Y$ between nodes $X$ and $Y$ denotes that $X$ is a direct cause of $Y$ with respect to the rest of the variables.

\begin{figure}[t]
\centering
\includegraphics[scale=0.35]{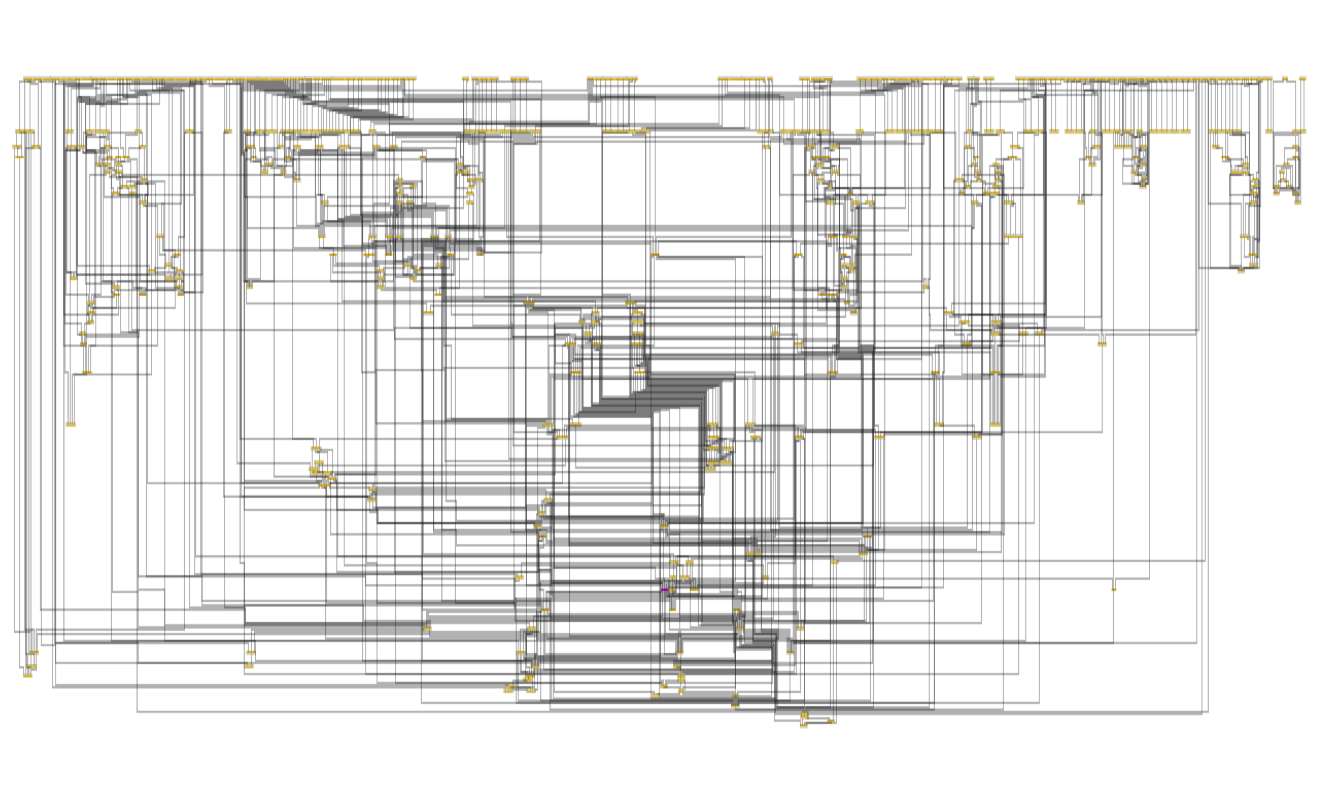}
\caption{Causal network learned by PC on real data using significance level $\alpha=0.05$ for the tests of conditional independence. The graph has 431 nodes and 1164 edges.
The resulting model is very dense and hard to interpret.}
\label{fig:vanilla-pc}
\end{figure}

\subsection{Learning causal models \\from observational data}

In this section we briefly review how to recover the structure of Bayesian networks from data.
Let $\mathbf{V} = \{V_1,\ldots, V_d\}$ be a set of $d$ random variables and $P(V_1, \ldots, V_d\}$ the joint distribution of these variables.
Let $\mathcal{D}=\{\mathbf{x_1}, \ldots, \mathbf{x_n}\}$ be a data set consisting of $n$ independent and identically distributed (i.i.d.) samples from the joint distribution over $\mathbf{V}$.
The task of causal discovery can be formalized as learning the structure of $G$ when given data set $\mathcal{D}$. 

Three main families of algorithms have been developed that learn the structure of causal models from data. 
\begin{enumerate}
    \item 
\emph{Constraint-based} algorithms operate in two phases. First, they use hypothesis tests to learn an undirected graph and then apply a series of orientation rules to retrieve a partially directed graph that corresponds to the Markov equivalence class of the true model. 
Algorithms in this category include PC~\cite{spirtes-etal-book00}, FCI~\cite{spirtes-etal-book00}, and Grow Shrink~\cite{margaritis2003learning}.
\item
\emph{Search-and-score} methods search heuristically through the space of possible directed acyclic graphs and pick the one that maximizes a scoring function. 
Such algorithms include greedy hill climbing search or search using tabu lists.
It is worth mentioning Greedy Equivalence Search~\cite{chickering2003optimal}, a score-based approach that searches over the space of equivalence classes, as opposed to the space of DAGs.
\item
\emph{Hybrid} algorithms combine elements from both of the aforementioned approaches.
They use hypothesis tests to limit the space of available models and then search over the constrained space to find the best oriented model.
An example of a hybrid algorithm is MMHC~\cite{tsamardinos2006mmhc}.
\end{enumerate}

In this work, we explore the use of constraint-based methods for learning the structure of causal models for manufacturing domains.

\subsubsection{The PC algorithm}

\begin{figure}
    \centering
    \includegraphics[width=0.8\columnwidth]{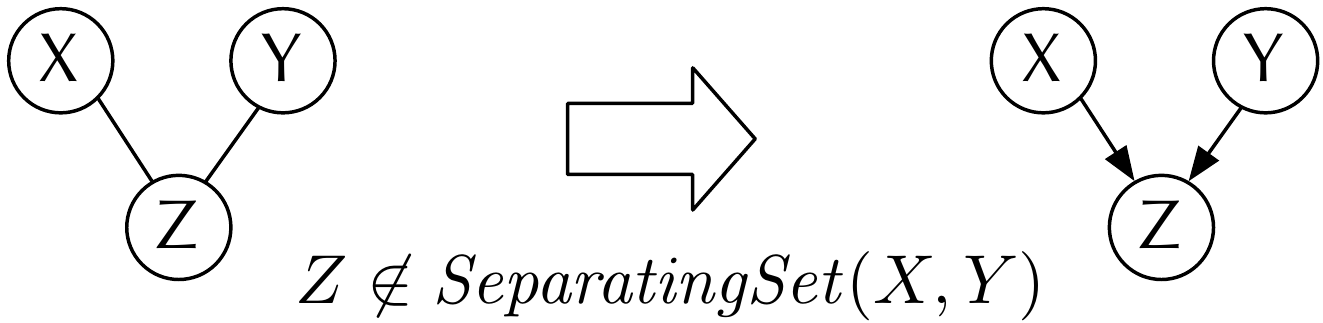}
    \caption{A schematic representation of how colliders are oriented. Every unshielded triple $X - Z - Y$ where there is no edge between $X$ and $Y$ is oriented as a collider if $Z$ DoEs not belong in the separating set recorded for $X$ and $Y$.}
    \label{fig:collider-detection}
\end{figure}

\begin{algorithm}[t]
\SetAlgoNoLine
\SetAlgoNoEnd
\LinesNumbered
\DontPrintSemicolon
\SetKwFunction{completeUndirectedGraph}{completeUndirectedGraph}
\SetKwFunction{applyOrientationRyles}{applyOrientationRyles}
\SetKwFunction{orientColliders}{orientColliders}

\nl$G\leftarrow \completeUndirectedGraph(V, E)$\;
\nl$S\leftarrow \{\}$\;
\tcp*[l]{Phase I}
\For{$d\leftarrow 0$ \KwTo $\mathit{depth}$}{
	\For{ undirected edge $ e=\langle X, Y\rangle\in \mathit{E}$}{
		\ForEach{$\mathit{condSet}\in \powerset(\mathit{Neighbors}[Y]\setminus\{X\})$}{
			\If{$\vert \mathit{condSet}\vert = d$}{
				\If{$X\indep Y\ \vert\ \mathit{condSet}$}{
				Remove $e$ from $E$\;
				$S[X, Y]\leftarrow \mathit{condSet}$\;
				\textbf{break}\;
				}
			}
		}
	}
}
\tcp*[l]{Phase II}
$G\leftarrow \orientColliders(G)$\;
\While{$changed$}{
	$G\leftarrow \applyOrientationRyles(G)$\;
}
$\Return\ G$
\caption{PC($\mathcal{D}$, $V$, $\mathit{depth}$)}
\label{alg:pc}
\end{algorithm}

In this work, we focus on constraint-based algorithms, and we specifically focus on the PC algorithm~\cite{spirtes-etal-book00}.
PC operates in two phases (pseudocode for the standard algorithm is shown in Algorithm~\ref{alg:pc}).
Phase I starts with a fully connected undirected graph and tests pairs of variables for conditional independence given conditioning sets of increasing size.
Once two variables $X$ and $Y$ are found to be independent given a separating set $\mathbf{S}$ (denoted as $X\indep Y|\mathbf{S}$), the edge between $X$ and $Y$ is removed and the separating set is recorded.
The output of Phase I is an undirected graph and a list of separating sets for each missing edge.
The absence of an edge between $X$ and $Y$ denotes that there exists a set of variables that render $X$ and $Y$ conditionally independent. 

Phase II of PC starts with the undirected graph produced by Phase I and aims to orient as many edges as possible. 
The first step is to orient the colliders (a schematic representation of how this is done is shown in Figure~\ref{fig:collider-detection}).
After all colliders have been oriented, a set of four orientation rules\footnotemark{} is applied repetitively until no changes can be made~\cite{meek1995causal}. 
The output of Phase II is a (partially) directed model that represents the Markov equivalence class of the underlying distribution. 
Under the assumptions of causal Markov condition, faithfulness, and causal sufficiency, PC has been shown to be sound and complete in the large sample limit (i.e., with perfect tests of conditional independence). 
This implies that the algorithm is guaranteed to find the a maximally oriented graph that is consistent with the independencies inferred from the data, under the aforementioned conditions.

\footnotetext{The four rules are: known-non-colliders, cycle avoidance, Meek's rule 3, and Meek's rule 4. The latter is necessary for the completeness of the PC algorithm in the presence of prior knowledge.}


\section{Methodology}

\begin{figure*}[t]
    \centering
    \begin{subfigure}{0.45\textwidth}
        \includegraphics[width=\textwidth]{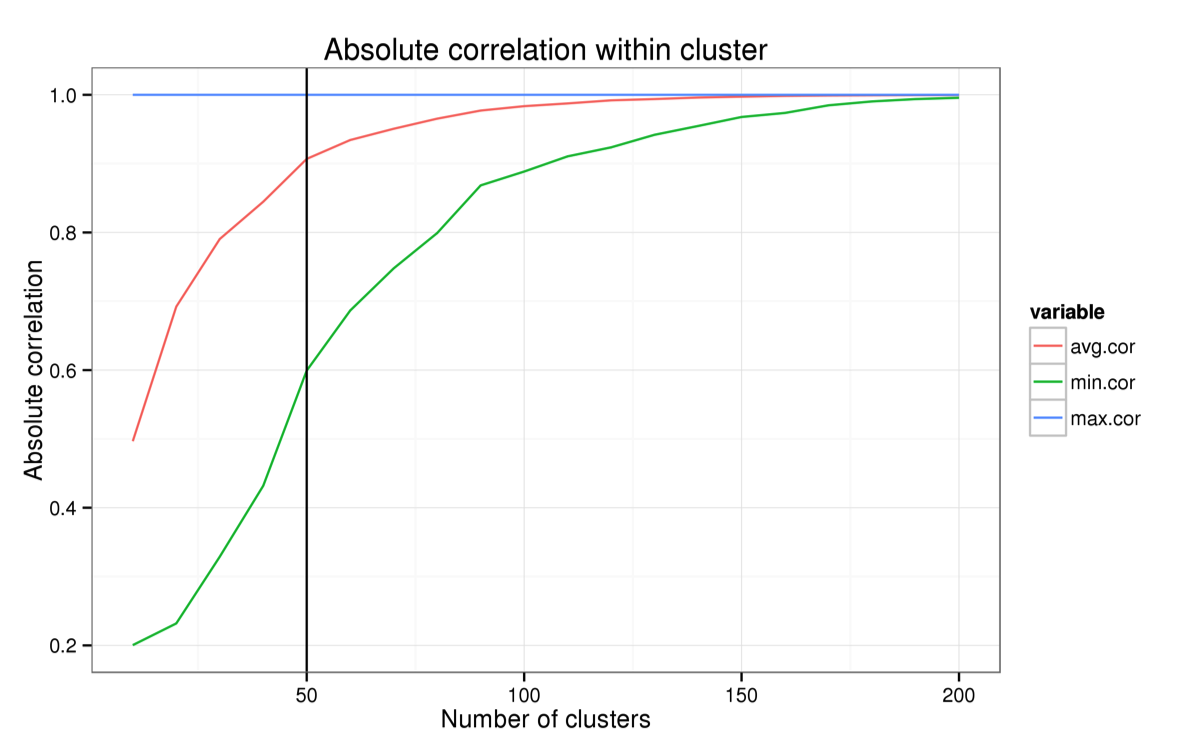}
        \caption{Within cluster correlation (minimum, maximum and average across clusters).}
        \label{subfig:avg-correlation}
    \end{subfigure}
    \qquad 
    \begin{subfigure}{0.45\textwidth}
        \includegraphics[width=\textwidth]{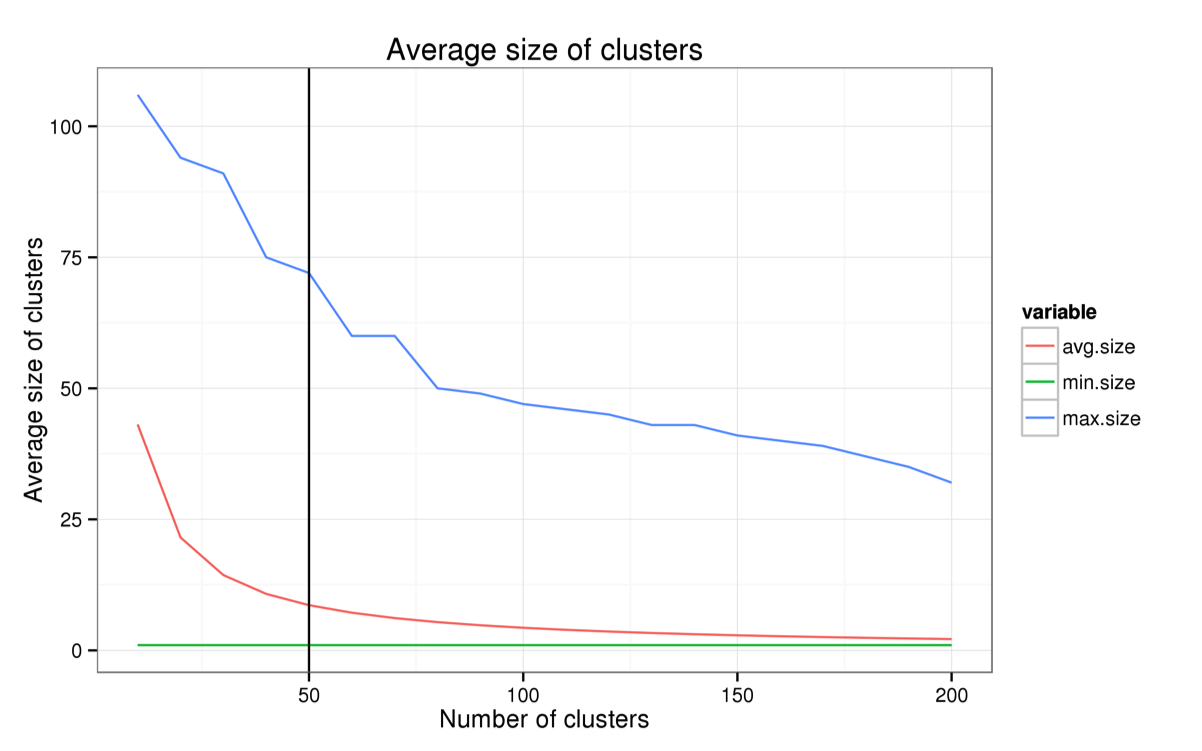}
        \caption{Cluster size (minimum, maximum, and average across clusters).}
        \label{subfig:avg-size}
    \end{subfigure}
    \caption{Clustering diagnostics for the features. For the clustering using 50 clusters, the average correlation within cluster is 80\% and the average size of clusters is around 10.}
    \label{fig:cluster-diagnostics}
\end{figure*}

First, to obtain a baseline, we applied the standard PC algorithm in the data using significance value $\alpha=0.05$ for the tests of conditional independence.
For the  PC algorithm, we used the implementation provided in the \texttt{pcalg} package~\cite{pcalg} (with appropriate modifications, as explained below). 
We used Fisher's z-transformation of the partial correlation as the conditional independence test (throughout the paper).

The learned graph is shown in Figure~\ref{fig:vanilla-pc}.
It consists of 431 nodes and 1164 edges. 
This simplistic approach demonstrates some of the challenges presented by manufacturing data. 
The learned model is hard for practitioners to interpret, both because of the large number of nodes and the density of the learned graphical model. 

We now present modifications on the PC algorithm that leverage the properties of manufacturing data to learn an interpretable causal model of the domain.
Specifically, we consider the following adjustments:
\begin{enumerate}
    \item Incorporating the prior domain knowledge regarding the temporal ordering of the stations in PC (discussed in~\ref{sec:prior}).
    \item Finetuning the parameters of the conditional independence tests in the algorithm (see~\ref{sec:finetuning}).
    \item Reducing the number of features to consider in the algorithm (see~\ref{sec:concept-detection}).
\end{enumerate}

\subsection{Incorporating Prior Knowledge in PC}
\label{sec:prior}

The first modification aims to leverage the inherent temporal constraints of a manufacturing process by incorporating prior knowledge in the algorithm. 
Specifically, events in a manufacturing process are strongly sequential. 
There is a total ordering among the assembly and testing stations.
For example, in Figure~\ref{fig:AssemblyLine}, station 1 precedes station 2. 
This induces a partial ordering on the variables measured across all stations. 
That is, all variables measured at station 1, precede all variables measured at station 2, Therefore, variables of station 2 cannot be causal for variables measured at preceding stations. 
However, among the variables measured in the same station, we have no information about their ordering. 

In order to improve the results, we incorporated the available prior knowledge (partial ordering of the variables) in PC. 
Specifically, after phase I of PC, where we have an unoriented graph, we oriented all edges for which we have temporal information. 
Then, we ran the second phase of PC to orient as many of the remaining edges as possible. 
This does not produce a more sparse network, but it does produce a more accurate one.

\subsection{Finetuning PC}
\label{sec:finetuning}

For the PC algorithm, there are two main parameters to adjust: the type of conditional independence tests used and the significance level ($\alpha$ value) for these tests. 
In this work, we only considered different values for the significance level. 
For the tests of conditional independence, if the computed statistic has a p-value lower than the specified significance threshold, the null hypothesis of independence is rejected, and thus the edge is kept in the model. 
The higher the $\alpha$ value, the more edges will be kept in the model.
In other words, decreasing the $\alpha$ value results in sparser models.

Given the large sample size, the algorithm can detect even very weak dependencies from data. 
However, in many cases, weak dependencies are of no interest to researchers. 
In order to account for the effect weak dependencies, we incorporated a strength-of-effect cutoff in the conditional independence tests.
We used the square of partial correlation to measure the strength of effect. 
Dependencies weaker than a specified threshold were ignored. 
Decreasing the strength of effect threshold results in denser models. 

\subsection{Clustering features}
\label{sec:concept-detection}

\begin{figure*}[t]
    \centering
    \begin{subfigure}[t]{0.4\textwidth}
    \includegraphics[width=\textwidth]{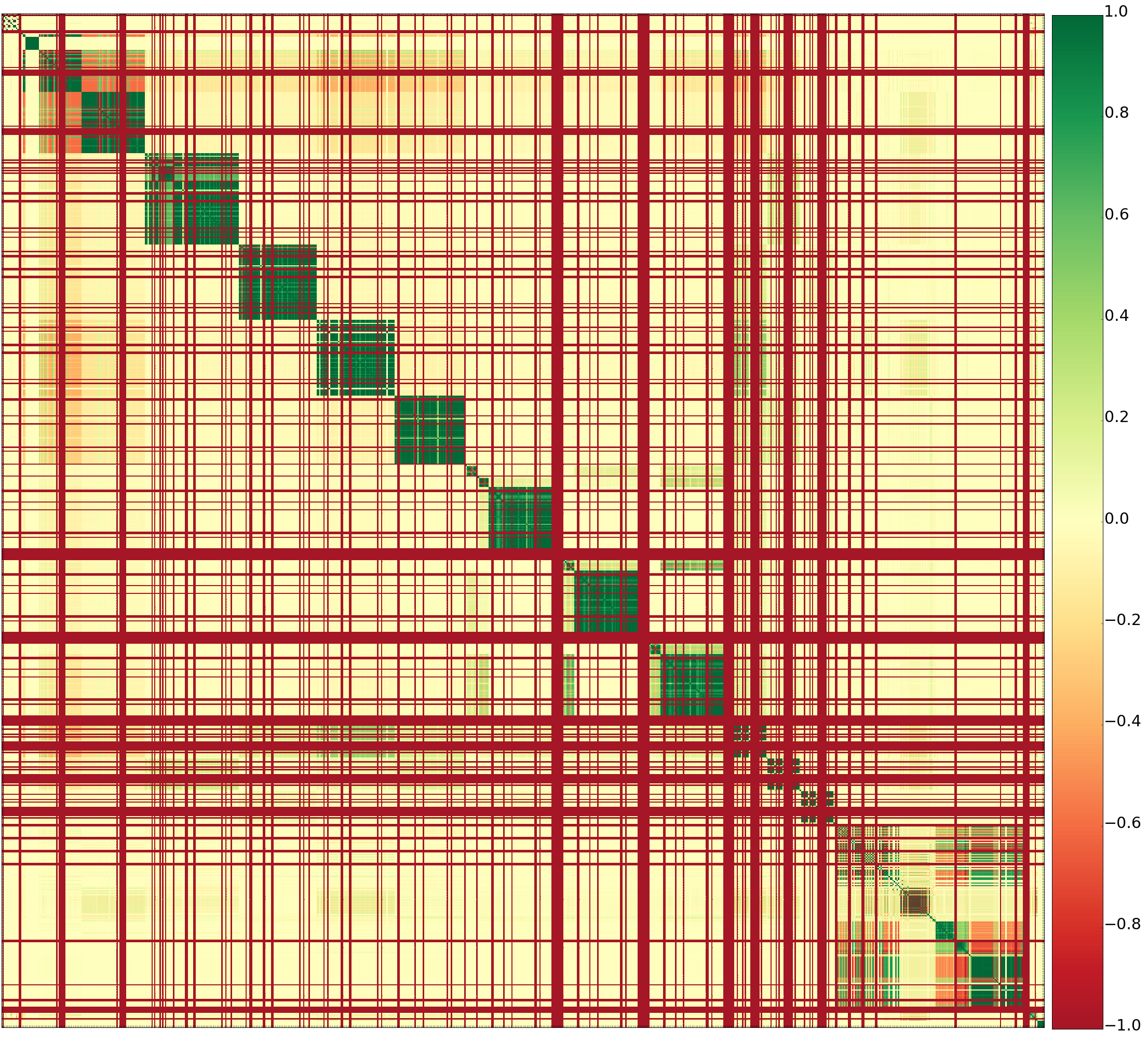}
    \caption{Pairwise correlation between the original features.}
    \label{fig:correlation-original}
\end{subfigure}
~
\begin{subfigure}[t]{0.4\textwidth}
    \centering
    \includegraphics[width=\textwidth]{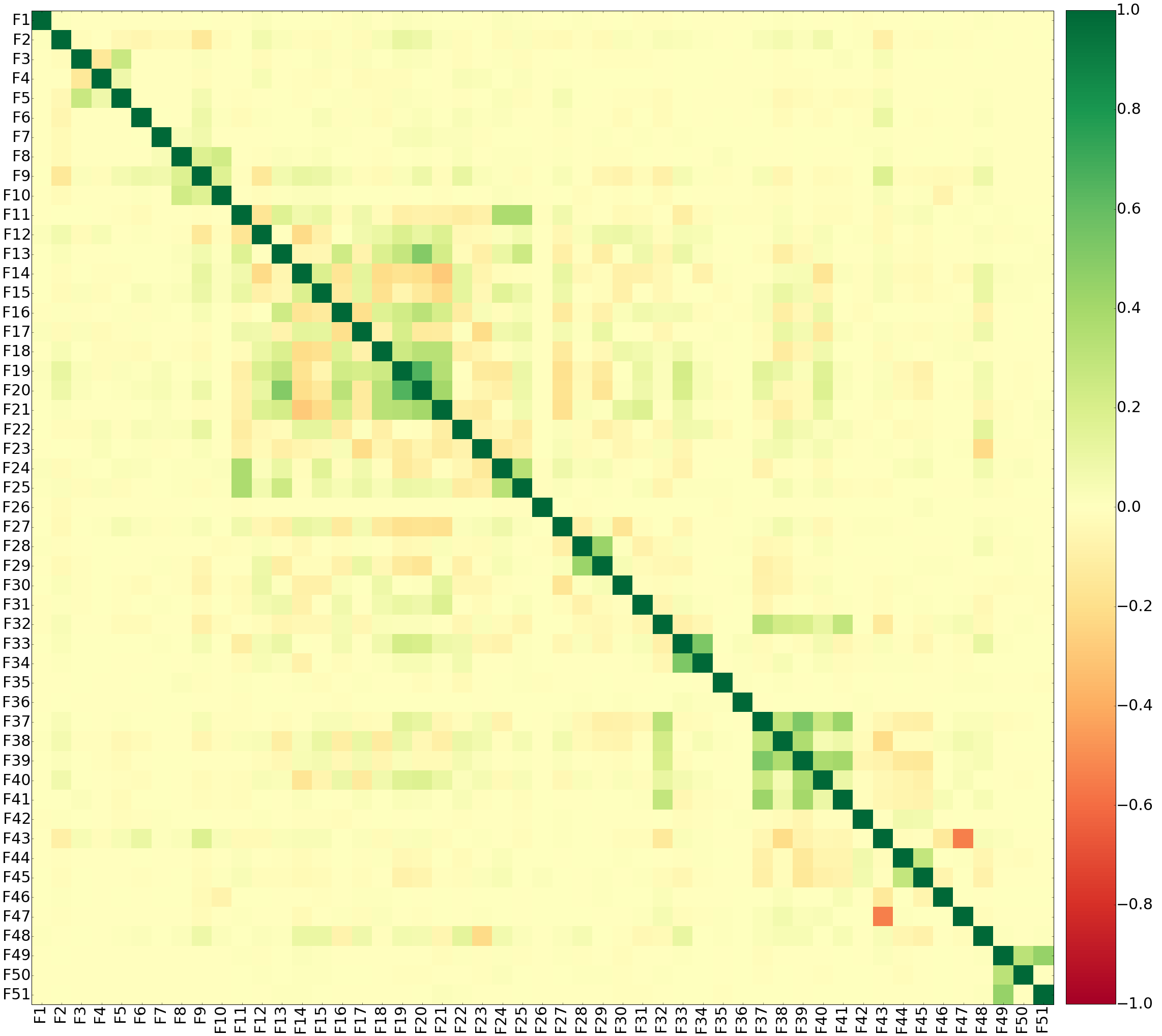}
    \caption{Pairwise correlation between the medoids of the clusters.}
    \label{fig:correlation-clustered}
\end{subfigure}
\caption{Pairwise correlation between features before (left) and after (right) applying hierarchical clustering.}
\label{fig:correlation}
\end{figure*}

The measurements taken along the assembly line exhibit high pairwise correlation, as shown in Figure~\ref{fig:correlation-original}.
This is especially true for measurements taken in the same station.
This has two consequences for structure learning algorithms. 
First, features that are highly correlated will remain connected in the learned model. 
This results in a dense graph and hinders interpretability.
Moreover, as noted above, the large number of features by itself is one of the factors that contribute to the uninterpretability of the learned causal model. 
The fact that certain features are almost perfectly correlated, provides a starting point for decreasing the number of variables.

In order to reduce the number of features to be included in the causal model, we clustered highly correlated features together using hierarchical clustering. 
The distance between two features $X$ and $Y$ was defined as 
\[
d(X, Y) = 1 - |\mathit{correlation}(X, Y)|.
\]
The distance between two clusters was computed as the maximum distance between their individual components (i.e., maximum linkage)\footnotemark{}.
From each cluster, we chose the most representative feature (the medoid) as the feature that exhibits the maximum average correlation with every other feature in the same cluster.
The resulting clusters are coherent, in the sense that each cluster contains mostly variables from the same station. 
\footnotetext{Other linkage options were also used (average, median, centroid, Ward) with similar results.}

To provide a qualitative evaluation for the quality of the resulting clustering, Figure~\ref{subfig:avg-correlation} shows the minimum, maximum, and average correlation across clusters as the number of clusters increases. 
Similarly, Figure~\ref{subfig:avg-size} shows how the size of the clusters varies when the number of clusters increases.
Moreover, Figure~\ref{fig:correlation-clustered} shows the pairwise correlation between the medoids of clusters.
In this case, the selected medoids exhibit less pairwise correlation compared to the original features.
Finally, the learned causal graph using only the cluster medoids is shown in Figure~\ref{fig:pc-medoids}.
This is a significantly smaller model, compared to that of Figure~\ref{fig:vanilla-pc}, and one that is easier to interpret.


\section{Evaluation}

So far, we have presented \emph{qualitative} results that show how clustering and domain knowledge can improve the interpretability of the learned causal model.
Below, we provide a quantitative evaluation of the impact that prior knowledge, parameter fine-tuning, and feature clustering have on causal discovery.

Ideally, the performance of a structure-learning algorithm would be evaluated through studying the effects of interventions. 
For example, to establish that $X$ is a cause of $Y$, we would intervene on $X$ and observe if the distribution of $Y$ changes in the way predicted by the model. 
In practice, this type of direct interventions in an assembly line comes at a high cost. 
To evaluate the quality of the models learned through PC for the manufacturing process, we used two proxies for real interventions:
\begin{enumerate}
    \item We compared the learned model against certain ``true causal relations'', as identified by domain experts. 
    This establishes the usefulness of causal discovery methods for manufacturing domains. 
    \item In order to evaluate different variants of the PC algorithm, we used synthetic models to generate data similar to the real data produced by the manufacturing line. 
    We then compared the accuracy of models learned through different variants of PC against the generative model. 
\end{enumerate}

\subsection{Evaluation through domain expertise}

\begin{figure}[t]   
\centering
    \begin{subfigure}{0.5\textwidth}
        \includegraphics[width=\columnwidth]{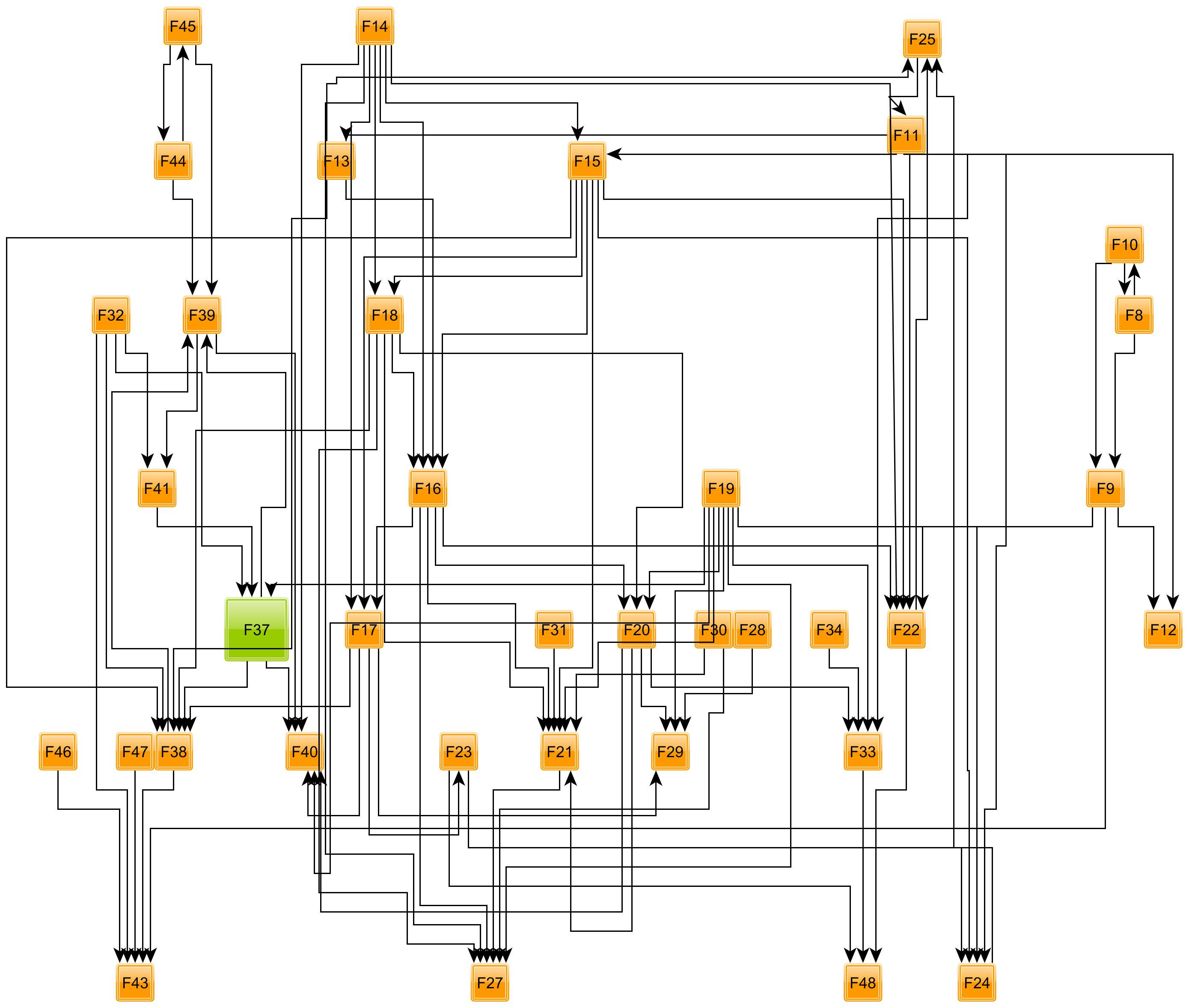}
        \caption{Causal model learned by PC using only the medoids of the 50 clusters created with hierarchical clustering. The green node is the cluster that corresponds to the target variable (yield). Note that the feature names have been anonymized.}
        \label{fig:pc-medoids}
    \end{subfigure}
 \\   
    \begin{subfigure}{0.5\textwidth}
        \includegraphics[width=\columnwidth]{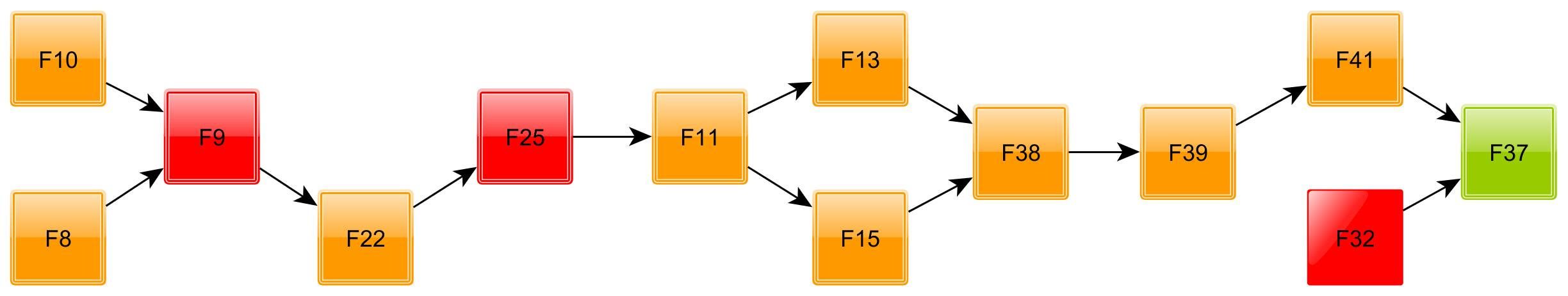}
        \caption{The paths containing critical features provided by the Domain experts (in red).}
        \label{fig:domainexpert-validation}
    \end{subfigure}
    \caption{Application of PC on the clustered features on real data.}
\end{figure}

One way to partially evaluate the effectiveness of causal structure learning in this manufacturing domain is through the use of domain knowledge. As noted earlier, Figure~\ref{fig:pc-medoids} contains causal relationships extracted by our model on a reduced feature set. Domain experts also provided partial ground truth where they identified nine critical features that are causal for the target variable of interest, based on their expertise and knowledge of the physical properties of the manufacturing line.
We indexed each of these features using the medoids of the clusters from the hierarchical clustering that they belonged to. We found that these nine features belonged to three clusters with medoids F32, F25 and F9. From Figure~\ref{fig:pc-medoids}, we observe that F32, F25 and F9 indeed are identified as causes of the target variable F37. Figure~\ref{fig:domainexpert-validation} contains the exact paths extracted from the full causal model that contain these critical features. This confirms that the causal structure learned from the data agrees with the causal paths provided by the domain experts. 


\subsection{Evaluation through the use of\\ synthetic models}

The fact that the learned causal model matches the intuition and knowledge of domain experts is very encouraging. 
However, it only provides validation for a small part of the model (the nine features identified as causes for the target variable). 
Unfortunately, the lack of ground truth makes the evaluation of the complete causal model impossible (at least without performing experiments directly on the production line). 
To quantify the performance of causal discovery techniques in the manufacturing domain given the absence of ground truth, we turn to synthetic models and simulated data. 
Specifically, we use a synthetic model to generate data and we treat the generating model as the ground truth. 
Ideally, the synthetic model will be similar to the actual model that describes the domain, and the generated data will resemble the real data produced by an assembly line. 


\subsubsection{Generation of synthetic data}

The first step towards the evaluation through synthetic data is the generation of the synthetic model and data.
The procedure we followed is outlined below:
\begin{itemize}
    \item We used a data set $\mathcal{D}_{\mathit{original}}$ containing data produced by an actual production line to learn a causal model $\mathcal{M}_{\mathit{true\_ec}}$ using the PC algorithm with parameters $\alpha=0.05$ and strength of effect equal to 0.01. 
    Note that we used the available temporal information when learning $\mathcal{M}_{\mathit{true\_eq}}$.
    $\mathcal{M}_{\mathit{true\_eq}}$ represents a Markov equivalence class and thus, contains both directed and undirected edges. 
    To generate data, we need a fully directed model.
    Therefore, we randomly picked a member of the Markov equivalence class $\mathcal{M}_{\mathit{generating}}$.
    This is a fully directed model that, by definition, respects the orientations enforced by the prior knowledge. 
    \item We then learned parameters for $\mathcal{M}_{\mathit{generating}}$ using the original data set $\mathcal{D}_{\mathit{original}}$. 
    The parametric form of the conditional distributions was assumed to be Gaussian.
    \item We used the learned model $\mathcal{M}_{\mathit{generating}}$ with the learned parameters to generate a set of 10 synthetic data sets, $\mathcal{D}_{\mathit{synthetic}}^i$, $i=\{1, \ldots, 10\}$ with 50000 data points each\footnotemark.
    \footnotetext{To learn the parameters of the Bayesian network and generate data from the model, we used the \texttt{bnlearn} package for \texttt{R}~\cite{bnlearn}.}
    \item Finally, we ran the PC algorithm on the synthetic data using varying values for the significance level and the strength of effect:
    \begin{align*}
        &\alpha =\{0.001, 0.01, 0.05, 0.1\}\\
        &\mathit{soe} = \{0, 0.05, 0.1\}
    \end{align*}
\end{itemize}

\begin{figure}[t]   
\centering
    \begin{subfigure}{0.4\textwidth}
        \includegraphics[width=\columnwidth]{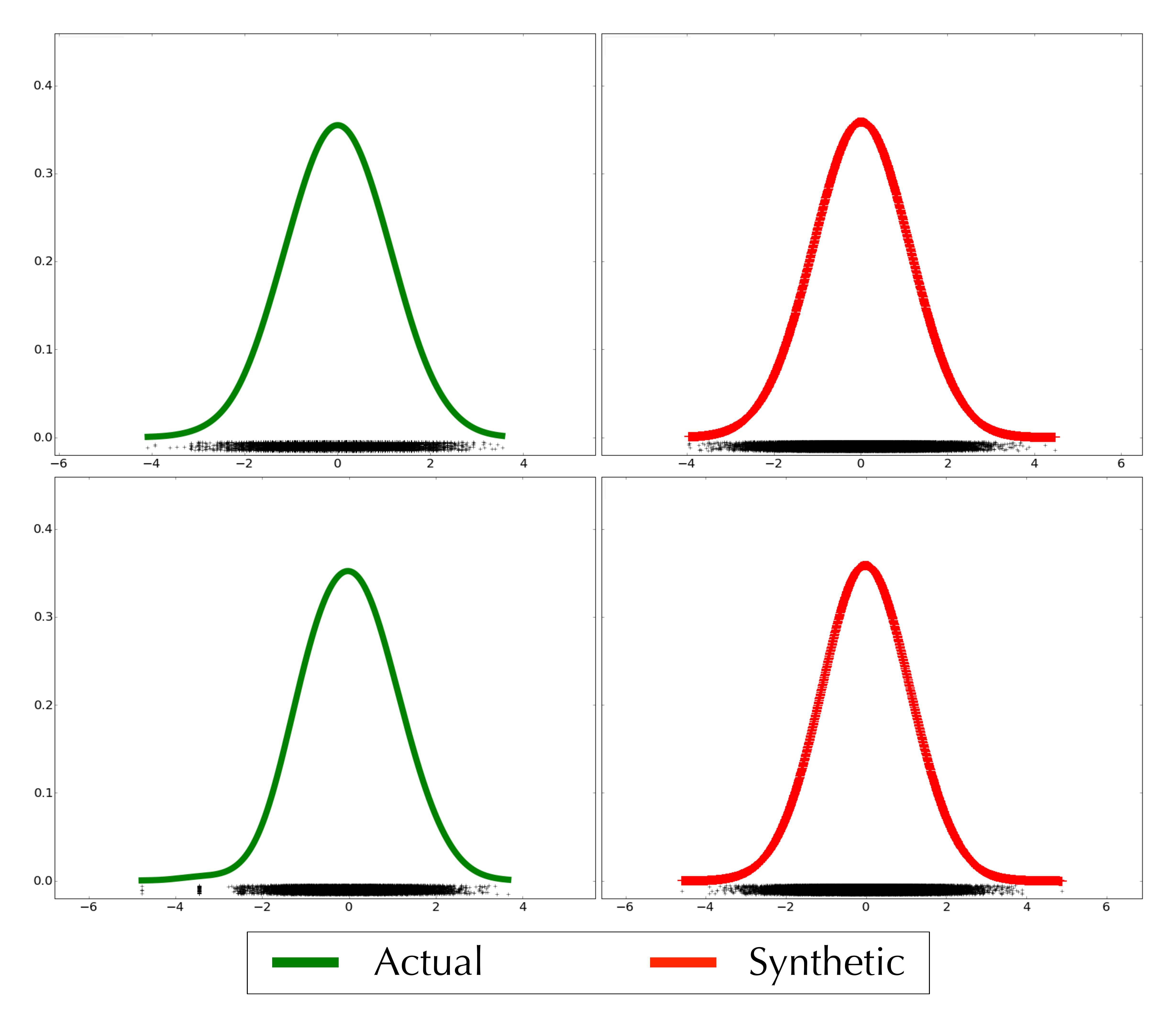}
        \caption{Example features where the actual data follows a normal distribution.}
        \label{fig:good-example}
    \end{subfigure}
 \\   
    \begin{subfigure}{0.4\textwidth}
        \includegraphics[width=\columnwidth]{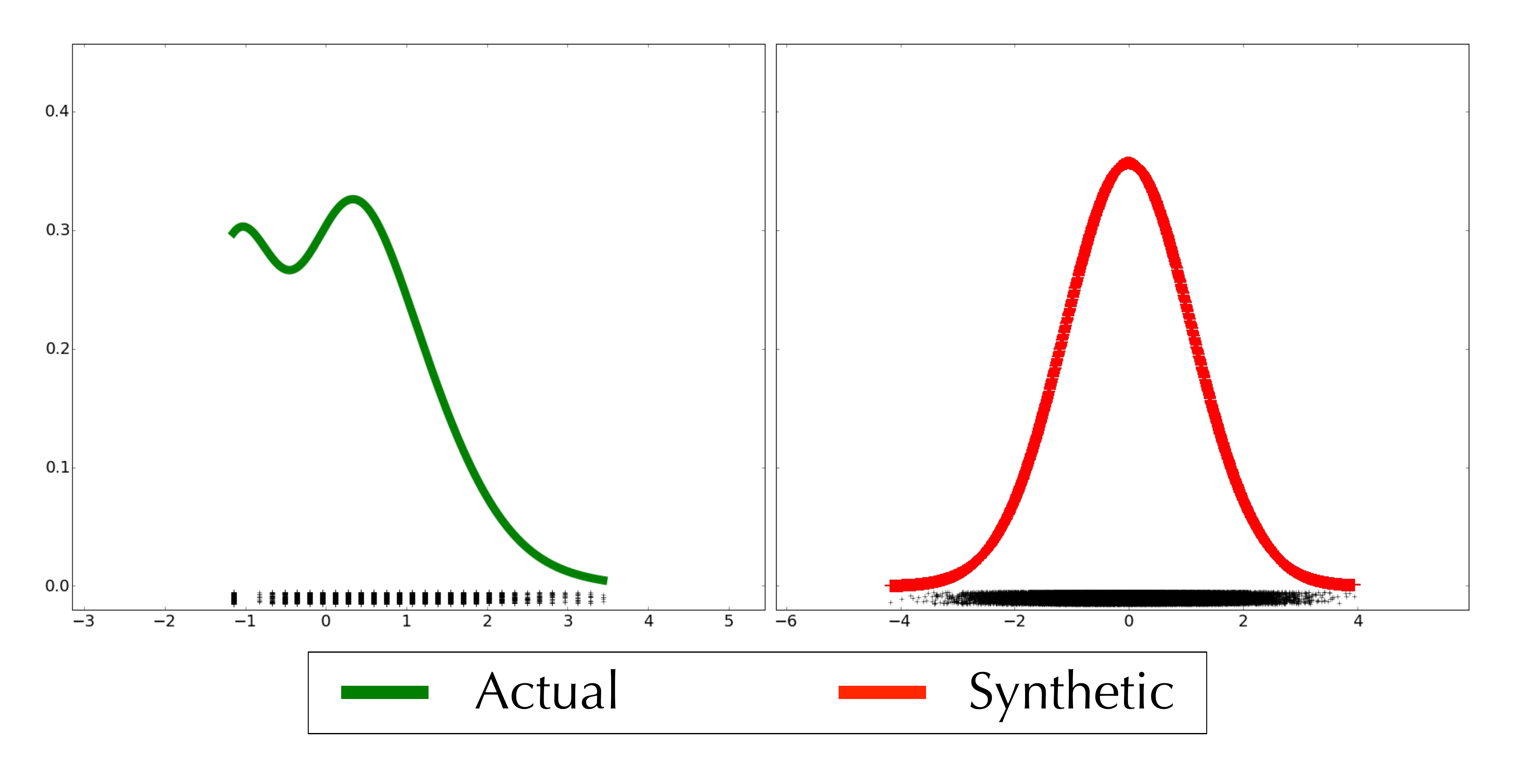}
        \caption{Example feature where the actual data follows a non-normal distribution.}
        \label{fig:bad-example}
    \end{subfigure}
    \caption{Comparison between the density of actual features and their synthetically generated counterparts.}
    \label{fig:actualVsSynthetic}
\end{figure}

The process we followed to generate synthetic data assumes that the features follow a Gaussian distribution. We evaluated this assumption by calculating the \emph{skewness} and \emph{kurtosis} for each feature distribution. Out of the 50 medoids we have extracted, we observed 26 features to have \emph{skewness} and \emph{kurtosis} values within the range of -2 to +2. This empirically indicates that 52\% of the features are within an acceptable range that indicates normality.
We compared the distribution of the generated features to that of the actual features, in order to demonstrate that the synthetic data we produce resemble the real data. 
Figure~\ref{fig:actualVsSynthetic} depicts the distribution of the actual feature compared to that of the generated features for three features.

\subsubsection{Results on synthetic data}

To evaluate the performance of PC on synthetic data we used precision and recall after Phase I (undirected model) and after Phase II (partially directed model). 
\begin{align*}
&\mathit{UndirectedPrecision}=\frac{\text{\# of true learned edges}}{\text{\# of learned edges}}\\
&\mathit{UndirectedRecall}=\frac{\text{\# of true learned edges}}{\text{\# of true edges}}\\
&\mathit{DirectedPrecision}=\frac{\text{\# of true learned oriented edges}}{\text{\# of learned oriented edges}}\\
&\mathit{DirectedRecall}=\frac{\text{\# of true learned oriented edges}}{\text{\# of true oriented edges}}
\end{align*}

\begin{figure}[t]
    \centering
    \includegraphics[width=0.8\columnwidth]{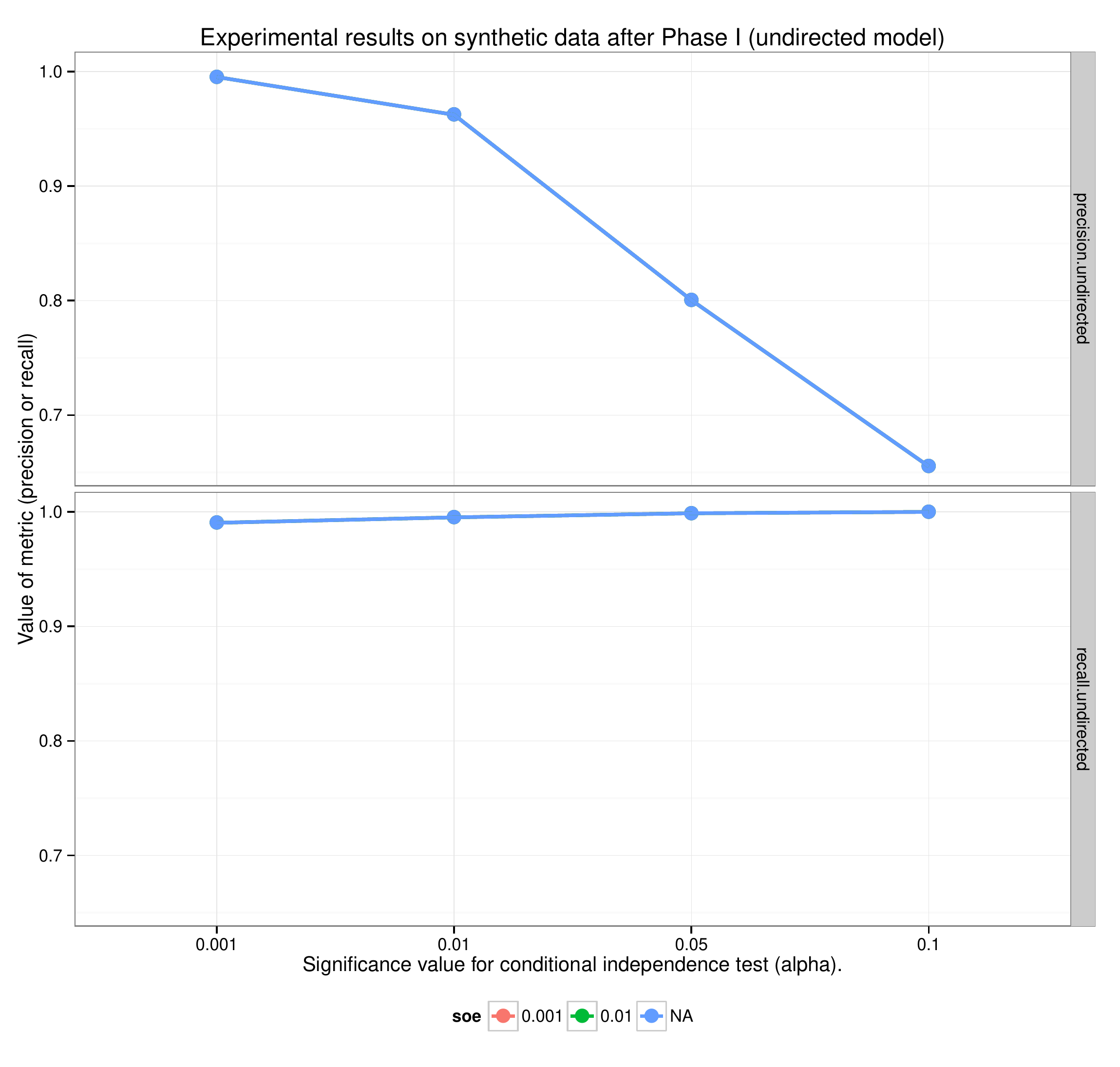}
    \caption{Average precision and recall after Phase I across the 10 synthetic data sets for varying parameters of PC.
    The generating model was learned with $\alpha=0.05, \mathit{soe}=0.01$.
    Note that the y-axis range starts at 0.65.}
    \label{fig:pr-undirected}
\end{figure}

\begin{figure}[t]
    \centering
    \includegraphics[width=0.8\columnwidth]{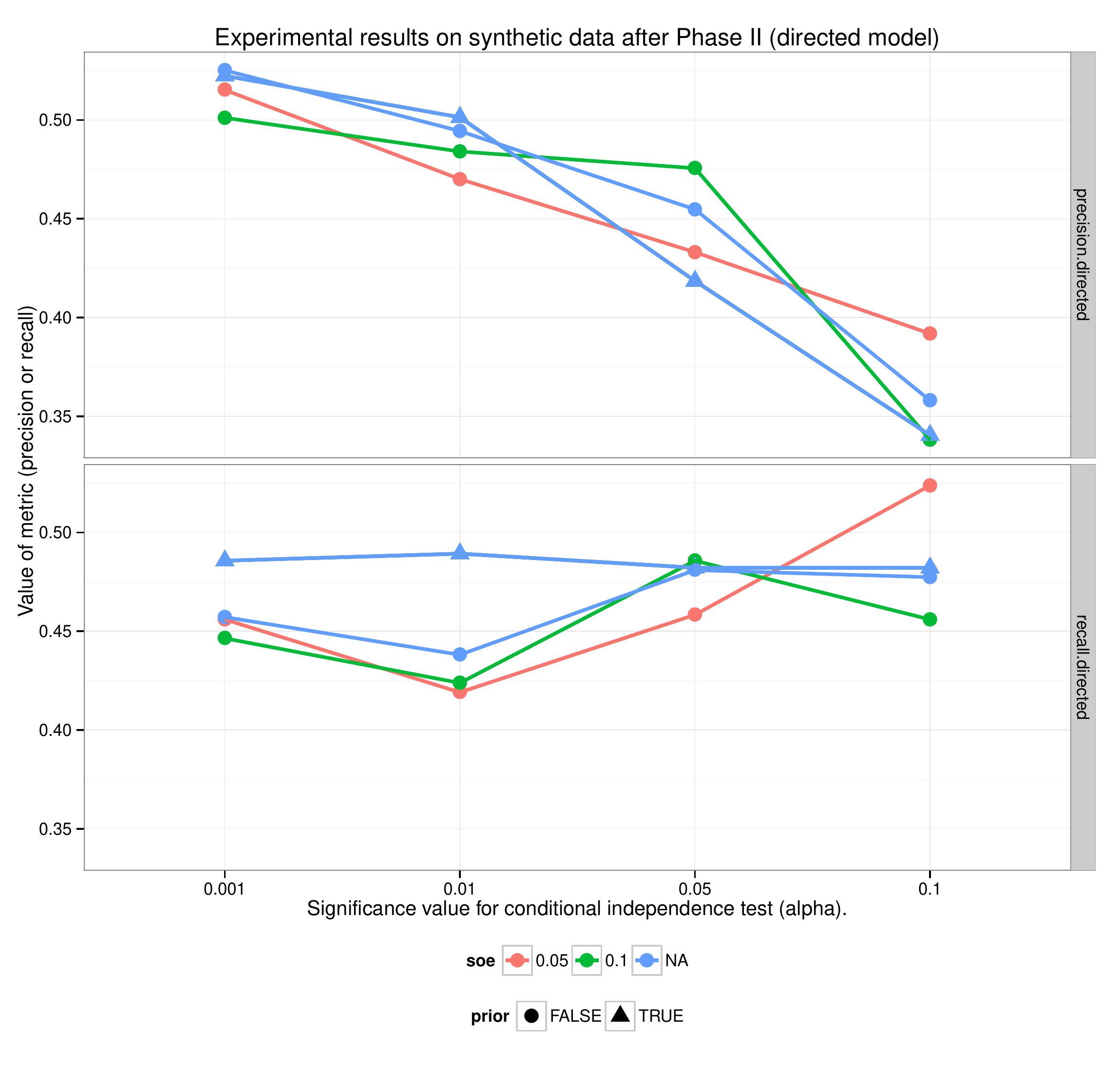}
    \caption{Average precision and recall after Phase II across the 10 synthetic data sets for varying parameters of PC. The generating model was learned with $\alpha=0.05, \mathit{soe}=0.01$.}
    \label{fig:pr-directed}
\end{figure}

The average precision and recall values across the 10 synthetic data sets after Phase I are shown in Figure~\ref{fig:pr-undirected}.
The algorithm has almost perfect recall (it learns all the true edges). 
The precision of the algorithm drops as the significance level of the test increases. 
This is to be expected, because higher $\alpha$ values result in a denser graph, therefore, more edges will be incorrectly included in the output. 
The results after Phase II are presented in Figure~\ref{fig:pr-directed}.
As the alpha values increases, the recall increases and the algorithm retrieves more than 50\% of the true directions.
However, the precision drops, suggesting that it is making more mistakes in the orientations. 
This can be explained because the learned model includes more spurious (directed) edges.

\section{Related Work}


In the area of causal discovery, the constraint-based algorithms we focused on retrieve models up to the Markov equivalence class (and thus might contain undirected edges). 
The authors of~\cite{sun2008distinguishing, sun2006causal} describe principled ways to choose a specific model from the Markov equivalence class.
Moreover, apart from the structure learning algorithms that leverage conditional independence, there exist other techniques to retrieve causal relationships from observational data. 
In the past few years, a group of methods based on additive noise models (ANMs) has been proposed~\cite{shimizu2006linear, hoyer2009nonlinear, peters2011identifiability, peters2011causal}.
ANMs leverage properties of the joint distribution other than conditional independence.
In short, it has been shown that if the observational distribution can be modeled as a structural equation model with an additive noise structure, then, under certain conditions, the directionality of all edges becomes identifiable.

Another line of related work focuses on learning graphical models for groupings of variables. 
Segal et al.~\cite{segal2005modulenetworks} introduce module networks which construct groups of variables with ``similar behaviour'' (i.e., variables that share the same set of parents and the same parameters), called modules.
They also present an  algorithm to learn a dependency structure between modules from data. 
Moreover, recent work by Parviainen et al.~\cite{parviainen2015bayesian} discusses the use of grouped variables for the specific task of causal discovery.

Regarding the application of causal discovery techniques and Bayesian networks on domains similar to the manufacturing domain, 
Verron et al.~\cite{verron08c, verron10b} use variations of a naive Bayes classifier (simple naive Bayes, tree-augmented Bayes classifier) to detect errors in manufacturing processes. 
Their task is however distinct than ours, as they aim to classify/predict if a part is faulty, as opposed to building a joint causal model of the domain.
More closely related is the work of Li et al.~\cite{Li-2007-knowledgeDiscoveryProcessControl}, that applied causal discovery techniques for process control data.
The above paper focuses on domain expertise for feature selection.
Moreover, the authors discretize their data and use domain knowledge to limit the number of states for every discrete variable.
In our work, we use statistical methods for feature selection/construction from the abundant raw features.
Finally, Nannapaneni et al.~\cite{nannapaneni2015automated} use Bayesian networks for uncertainty quantification for manufacturing domains. 
However, they are using score-based methods for construction of Bayesian networks from data.

\section{Conclusions and Future Work}

In this work, we apply causal structure learning techniques to the field of manufacturing. 
We highlight the challenges and opportunities presented by manufacturing data and show how causal algorithms can be adapted to such data. This implies a large feature space and the variables measured might be highly correlated. These characteristics make the task of causal discovery challenging.
On the other hand, there exists a large amount of domain knowledge, including both temporal and physical properties of the assembly line as well as knowledge from domain experts, that can be leveraged to improve the results.
We presented results from the application of the PC algorithm on real data produced by manufacturing lines. We incorporated prior knowledge and clustered the feature space to improve the precision of the model. The structures learned by these models have been partially evaluated by the domain experts and thus can be used to identify influential factors effecting production yield. We have also explored the behavior of the algorithm on synthetic data generated to be similar to the original data. We show that, in the absence of a ground truth, we can use the synthetic data to evaluate the accuracy of different algorithms for learning causal structure.

Data produced by manufacturing lines provide many opportunities for future work.
One potential direction is to verify some of the learned causal relationships through experimentation in the production line.
This intervention in assembly-line operations to determine if the observed changes match those predicted by the model. 
The cost associated with this approach is potentially very high. 
The assembly line will have to stop for interventions to take place and only a few causal relationships could be tested in this manner.
However, it would provide very strong evidence about the validity of the causal relationships that have been learned from data. 
Another direction for future work is to use information-theoretic measures of dependence instead of linear correlation (for example, mutual information on discretized data) and more advanced tests of conditional independence (such as the kernel-based conditional independence test~\cite{zhang2011kci}).
This will allow the models to identify non-linear relationships in the data.
Finally, the partial ordering of variables that arises naturally in manufacturing domains provides an ideal setting for application of top-down (or recursive) causal discovery algorithms~\cite{yehezkel2009bayesian,xie2008recursive,cai2013sada}.


\bibliographystyle{abbrv}
\bibliography{causal-discovery-manufacturing}

\end{document}